\documentclass{llncs}
\usepackage{llncsdoc}
\usepackage{graphicx}
\usepackage{url}
\usepackage{amsmath}
\usepackage{amssymb}
\usepackage{graphicx}
\usepackage[utf8]{inputenc}
\usepackage{booktabs}
\usepackage[bf,width=0.66\textwidth]{caption}

\begin{document}

% \fi

% \begin{document}

\title{Unsupervised Inflection Generation Using Neural Language Modeling}

% \title{MorphoGen: Full Inflection Generation \\ 
% Using Recurrent Neural Networks}

\author{Octavia-Maria \c{S}ulea$^{1,2,3}$, Steve Young$^{4}$}

\authorrunning{O.M. \c{S}ulea et al.}

\institute{$^{1}$ University of Bucharest, Faculty of Mathematics and Computer Science, Romania\\
$^{2}$Human Language Technologies Research Center, University of Bucharest\\
$^{3}$GumGum, Santa Monica, California, USA\\
$^{4}$Pronto.AI, San Francisco, California, USA \\
mary.octavia@gmail.com, steve.young@pronto.ai}

\maketitle

\begin{abstract}

The use of Deep Neural Network architectures for Language Modeling has recently seen a tremendous increase in interest in the field of NLP with the advent of transfer learning and the shift in focus from rule-based and predictive models (supervised learning) to generative or \emph{unsupervised} models to solve the long-standing problems in NLP like Information Extraction or Question Answering. While this shift has worked greatly for languages lacking in inflectional morphology, such as English, challenges still arise when trying to build similar systems for morphologically-rich languages, since their individual words shift forms in context more often \cite{sulea16}. In this paper we investigate the extent to which these new \emph{unsupervised} or \emph{generative} techniques can serve to alleviate the type-token ratio disparity in morphologically rich languages. We apply an off-the-shelf neural language modeling library\cite{woolf} to the newly introduced \cite{sulea19} task of unsupervised inflection generation in the nominal domain of three morphologically rich languages: Romanian, German, and Finnish. We show that this neural language model architecture can successfully generate the full inflection table of nouns without needing any pre-training on large, wikipedia-sized corpora, as long as the model is shown enough inflection examples. In fact, our experiments show that pre-training hinders the generation performance.
% \Keywords{computational morphology, inflection learning, deep neural networks} 
\end{abstract}

\section{Introduction}

Modeling variability, or the certain pattern a variable follows in context, is one of the major goals of Statistical Machine Learning and Data Science. In Artificial Intelligence, the notion whose variability researchers attempt to account for or simulate is the assumed concept of human intelligence. Within Natural Language Processing (NLP), where the variable scrutinized is language, this goal can surface into many different tasks, depending at which level of the language the analysis is being carried out. When we \footnote{preprint version} focus our attention at the word level, as researchers within Computational Morphology have done, we notice a clear distinction between languages that have a high \emph{stem variance}, meaning that the same word (stem) will shift its form often in the context of different sentences, or low \emph{stem variance}, where a word will appear in the same form in most contexts. The ones with high stem variance, like Romanian, French, or German, have been dubbed languages with \emph{rich} inflectional morphology (\cite{dinu-ranlp11}, \cite{dinu-eacl12}, \cite{dinu-ranlp13v}), while languages with low variance, like English, are considered to have a \emph{poor} inflectional morphology. This distinction has made applying machine learning algorithms to NLP tasks like Named Entity Recognition for languages like English much easier than for languages like French or German \cite{suleaND16}, where a base form (uninflected, dictionary form) of a word can go through many phonological transformations (apophonies) when they merge with affixes (sub-word character clusters) denoting gender, person, number, case, definiteness to reflect the context (inflection) they are being used in. While great progress in text classification, information extraction, or language understanding tasks has been made through language modeling with deep neural networks and transfer learning \cite{gpt}, \cite{gpt2}, these models still suffer when the background information needed for the task is out of vocabulary or rarely occurring and require fine-tuning on labeled datasets \cite{ulmfit}.

Tables \ref{tab:0} and \ref{tab:1} show the stem alternations occurring in Romanian and German nouns respectively as opposed to English nouns which incur no shifts in the stem. The ending of the stem and beginning of the contextual marker (inflectional affix) is marked with a hyphen. Note that although Romanian and German both are rich in inflectional morphology they do not share the same contexts for which alternation occurs. Specifically, Romanian nouns do not sift from Nominative to Accusative or from Genitive to Dative, unlike German, but they do shift depending on whether they are being marked for definiteness, a notion that in English is marked by the presence of determiners (i.e. \emph{the}).

\begin{table}[!h]
\centering
\caption{Stem alternations in the nominal domain for Romanian vs. English}
\label{tab:0}
\begin{tabular}{lll}
      \hline
		Tag & 	English & Romanian   \\
\hline

N-Acc.sg.indef   & (a) door-          & p\emph{oa}r\emph{t}-ă      \\
N-Acc.pl.indef   & (a few) door-s     & p\emph{o}r\emph{ţ}-i       \\
N-Acc.sg.def    & (the) door-         & p\emph{oa}r\emph{t}-a       \\
N-Acc.pl.def    & (the) door-         & p\emph{o}r\emph{ţ}-ii      \\
G-D.sg.indef    & (a) door-'s         & p\emph{o}r\emph{ţ}-i         \\
G-D.pl.indef    & (a few) door-s'     & p\emph{o}r\emph{ţ}-i        \\
G-D.sg.def      & (the) door-'s       & p\emph{o}r\emph{ţ}-ile      \\
G-D.pl.def      & (the) door-s'       & p\emph{o}r\emph{ţ}-ilor    \\ 

      \hline
\end{tabular}
\end{table}

\begin{table}[!h]
\centering
\caption{Stem alternations in the nominal domain for German vs. English}
\label{tab:1}
\begin{tabular}{lll}
      \hline

Tag & English & German\\

\hline

N.sg      & power     & m\emph{a}cht- \\
N.pl      & power     & m\emph{ä}cht-e\\
G.sg      & power     & m\emph{a}cht-\\
G.pl      & power     & m\emph{ä}cht-e\\
D.sg      & power     & m\emph{a}cht-  \\
D.pl      & power     & m\emph{ä}cht-en \\
Acc.sg.   & powers    & m\emph{a}cht-  \\
Acc.pl    & powers    & m\emph{ä}cht-e \\ 

      \hline
\end{tabular}
\end{table}

Over the last 10 years, there have been extensive efforts within the field of NLP to apply supervised learning to predict which inflectional class (morphological paradigm) words fall into: \cite{dinu-ranlp11}, \cite{dinu-eacl12}, \cite{dinu-ranlp13v}, \cite{durett13}, \cite{ahlberg15}, \cite{sulea16}. However, attempting to generate all inflected forms of a word from its base form in a fully unsupervised manner, given no explicit information of the existing paradigms of a language, has only very recently been studied for the verbal domain of a few morphologically rich languages \cite{sulea19}. We would like to pursue this investigation further and see to what extent it works in the more variable nominal domain. Our interest in the task of unsupervised inflection generation introduced in \cite{sulea19} was sparked by the application of neural generative models to the related task of morphological reinflection (\cite{farqui}, \cite{yoav17}, \cite{zhou_vae}) and by the recent availability of morphological datasets coming from Wikitionary \cite{durett13} and the SIGMORPHON shared tasks \cite{cotterell2016sigmorphon}. 

We make use of the same attention-based recurrent neural architecture \cite{woolf} applied in \cite{sulea19}, which has become a popular plug-and-play text generation tool outside of the academic community \footnote{airweirdness.com has used it extensively} and investigate the best practices and limitations to generating the paradigm of a noun given only its uninflected (dictionary) form and being trained in a fully unsupervised fashion. While \cite{sulea19} report state-of-the-art results for Romanian verb generation and show their models to require no fine-tuning, we show that in fact fine-tuning hinders the ability of the model to generate any inflection list. This comes as an argument against the pre-trained and fine-tuned strategy advocated in the past few years by the transfer learning community within NLP (\cite{ulmfit}, \cite{gpt}, \cite{gpt2}). We invite the NLP community to further test other neural models with this training scheme.

\section{Datasets}
For our experiments, we used a subset of the wikitionary corpus \cite{durett13} and the dataset subset of the corpus introduced in \cite{barbu2} and used in \cite{sulea16}, focusing on nominal inflection generation in German, Romanian, and Finnish. We chose the first two languages because they are known to display the morpho-phonlogical phenomenon of apophony, or stem alternations, and also represent two different language families: the Latin and Germanic. We also decided to look at Finnish since it is a Fino-Ugric language highly unrelated to the other two and, while it does have a rich inflectional morphology, it displays a characteristic that is arguably the opposite of apophony - agglutination (the stem does not alter during inflectional affixation).

While the German and Finnish datasets contained no direct (labeled) information about inflectional class, for Romanian we were able to attain the labeled dataset introduced in \cite{sulea16} which associates the dictionary form of a noun (nominative-accusative singular indefinite form) with a number from 0 to 20 representing the inflectional class label which they identified based on various linguistic works (\cite{barbu}, \cite{romalo}) and which was successfully predicted by their proposed model using only character ngrams of the base form. This label is supposed to encode the pattern of inflectional endings the noun receives as well as the pattern of alternations the stem goes through during inflection. The experiments in \cite{sulea19} show that making use of this label (by conditioning the language model) leads to lower generation performance so we did not carry this line of research over to the nominal domain.

\section{TextGenRNN Architecture}
We adapt the character-level language model \texttt{textgenrnn} developed in
\cite{woolf,felbo2017}, whose architecture is reviewed here for completeness.

% TODO: add description of embedding algorithm
Input sequences to the model are strings of up to $T$ characters. Each character
in an input sequence is first translated into a 100-D embedding-vector. These
are then fed through two bi-directional 128-unit LSTM layers. Next, the outputs
of the embedding and both LSTM layers are concatenated and fed into an attention
layer which weights the most important temporal features and averages them
together. Note that this skip-connects the embedding and first LSTM layer to the
attention layer, which helps alleviate vanishing gradients. Finally, the output
of the attention layer is routed through a fully-connected MLP layer with output
dimension equal to the number of possible characters.

% TODO: how does it generate?

\begin{figure}[t!]
  \begin{center}
    \includegraphics[angle=0,width=0.8\textwidth]{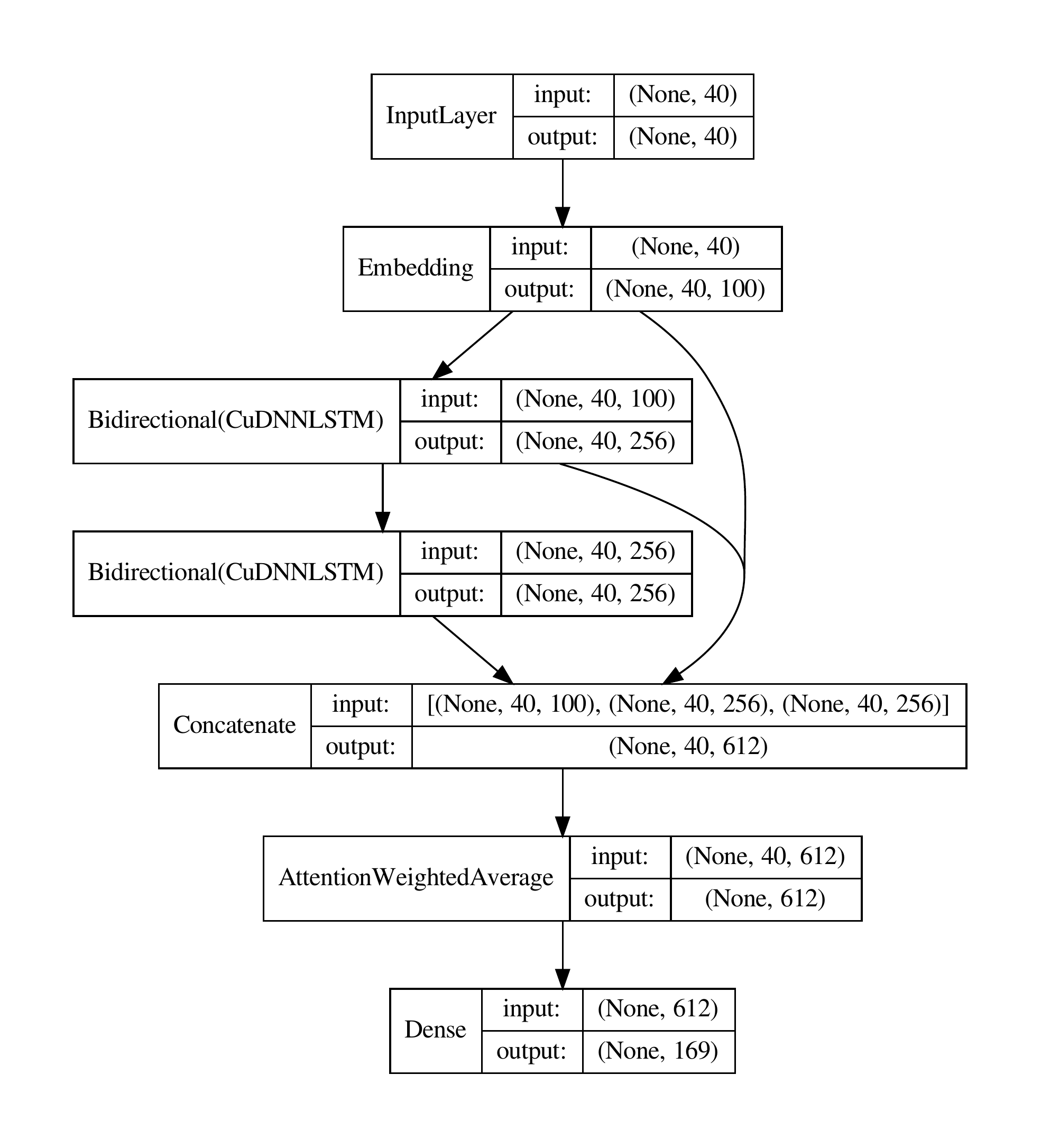}
    \caption{Structure of the model. We use a max-length $T=40$, a 100-D
      embedding matrix, 2 128-unit LSTMs, and an attention weighting with 169
      outputs. \label{the_model}}
  \end{center}
\end{figure}

\section{Generation Experiments}

The first set of experiments investigated the generation of full inflection given the dictionary form of the noun. Specifically, we wanted to see if the model could generate all forms of a noun 
% or 8 forms of a noun 
in one go, given only the uninflected (dictionary) form. For Romanian this would be the nominative-accusative singular indefinite form (N-Acc.sg.indef in Table \ref{tab:0}), whereas for German and Finnish this would be the nominative singular. To this end, the training datasets were arranged such that each row contained this base form of the noun, which is the first form to appear in dictionaries, followed by all the inflected forms resulted from combining case (nominative, genitive, accusative, and dative for German; nominative-accusative, genitive-dative for Romanian; 14 for Finnish), number (i.e. singular, plural) and definiteness.

% For nouns, this meant the 8 forms resulting from combining number with definiteness (i.e. indefinite, definite) and case (i.e. nominative-accusative, genitive-dative). 
Each form in the training set was separated by a comma and a space and written on the same line. The test sets were formatted the same. For German and Finnish, we consolidated the training and development sets together and used the test set for generation. For the Romanian set, since we also had access to the inflectional class for each base form, we split the 30k examples in train and test sets, making sure that both sets maintained the same distribution based on this inflectional class.

During the generation phase, the trained language model is given the uninflected form from the test set as the value for the \textit{prefix} parameter in the \textit{generate} function of the textgenrnn model. If the forms generated by the model match exactly with the test set forms, then the corresponding entry in the test set is counted as having been correctly generated. 

\begin{figure}[!h]
  \begin{center}
	\includegraphics[width=0.8\textwidth]{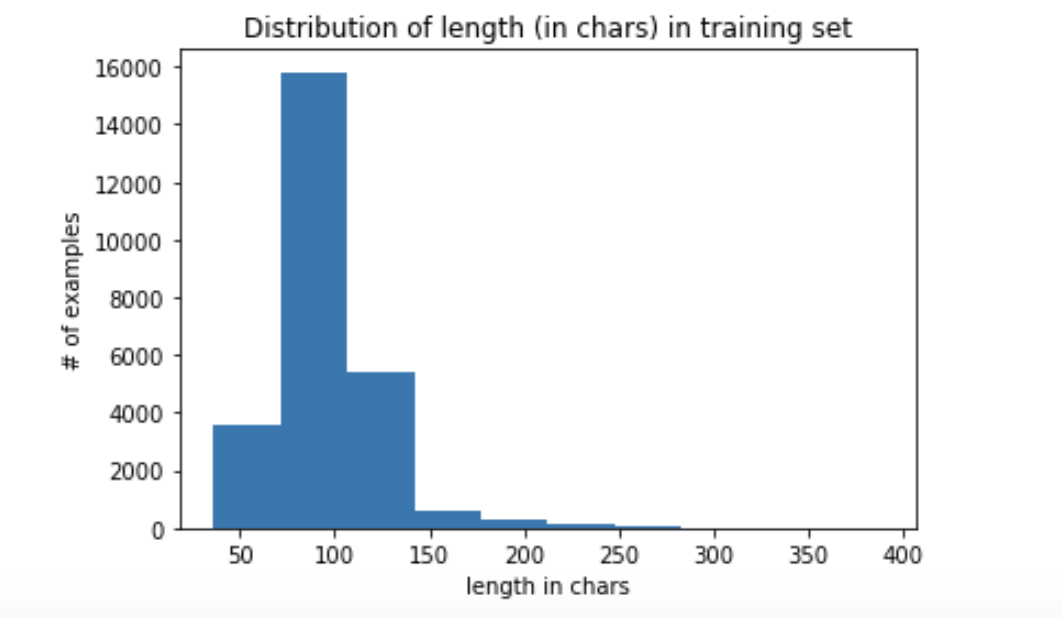}
	\caption{Distribution of character lengths for the Romanian training corpus}
	\label{fig2}
  \end{center}
\end{figure}

\begin{figure}[!h]
  \begin{center}
  \includegraphics[width=0.8\textwidth]{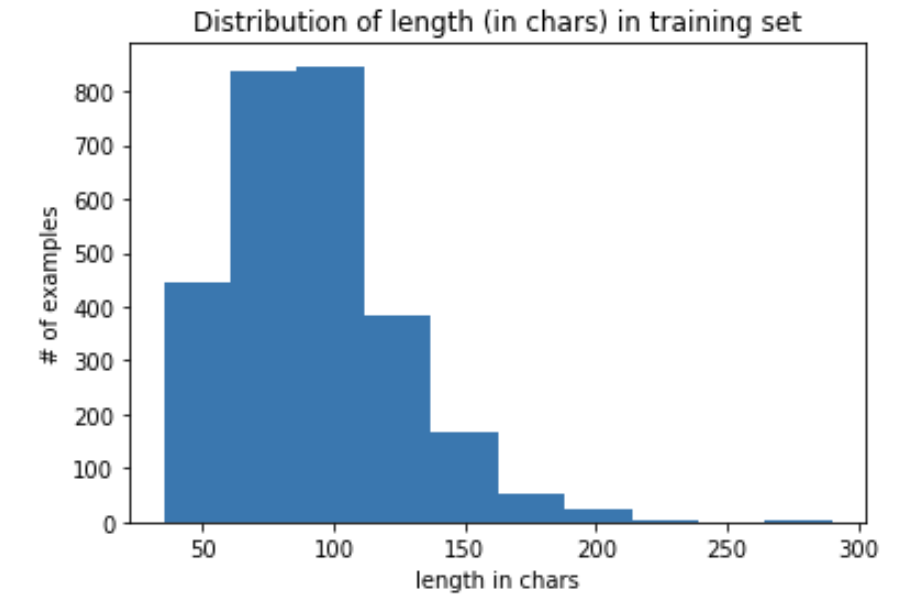}
  \caption{Distribution of character lengths for the German training corpus}
  \label{fig3}
  \end{center}
\end{figure}

The experiments were carried out with different training parameters. We trained the three language models for 14 epochs and different values for the \textit{max\_length} parameter which determines how many tokens (characters) are taken as context during training. Since the Romanian models performed less accurately than the German and Finish ones in the nominal domain but also than Spanish and Finnish in the verbal domain according to our previous study \cite{sulea19}, we investigated if pre-training of the model with the Romanian EuroParl corpus \cite{europarl} and fine-tuning it over the morphological corpus helped boost generation accuracy for the verbal model. We also tested the influence of the number of tokens to consider before predicting the next one by changing \textit{max\_length} from $40$ characters (default value) to $90$, chosen based on the distribution of lengths in the training corpora for Romanian \ref{fig2} and German \ref{fig3}. Interestingly, the Finnish corpus had a much larger character length average than the other two languages since it has considerably more cases (14 vs. 4). We decided to set the max\_length parameter to 100 for Finnish.

\begin{figure}[!h]
  \begin{center}
	\includegraphics[width=0.8\textwidth]{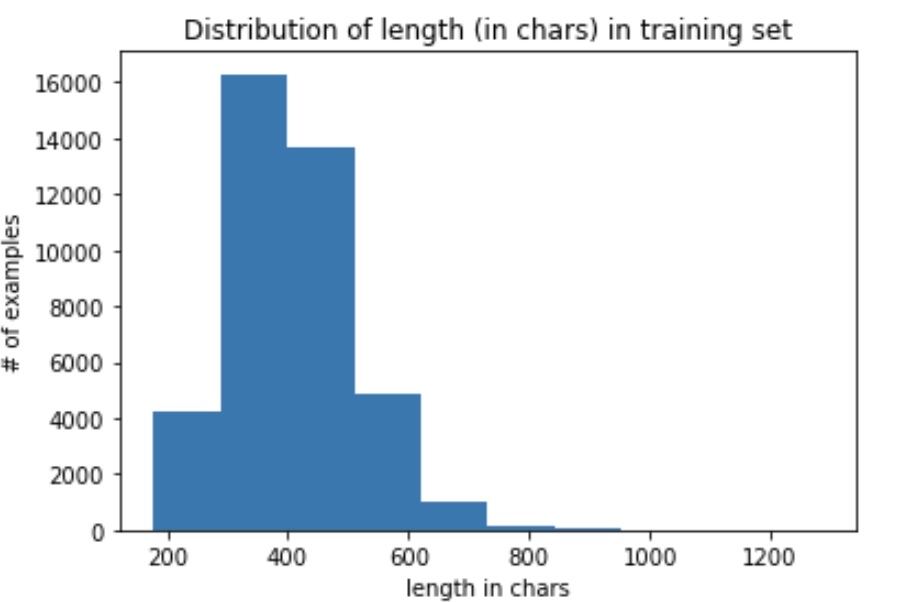}
	\caption{Distribution of character lengths for the Finnish training corpus}
	\label{fig4}
  \end{center}
\end{figure}

A final variable we investigated in our ablation experiments was the number of forms in each training example. Specifically, we wanted to see if, beside the number of characters we told the language model to look at when learning to predict the next character sequence, we also indirectly \emph{conditioned} the model to "understand" that it needs to generate a different amount of forms than the norm for certain words. Indeed, unlike the German and the Finnish datasets, whose examples consistently had 8 and 30 forms respectively, the Romanian noun dataset contained certain nouns which had both male and female variants and therefore the same base form would lead to more than 8 inflected forms. We tested the extent to which this architecture is able to learn, without explicit labeling of the dataset, so still in a fully unsupervised manner, the distinction between those nouns who get more than 8 forms by training two separate version of the model: one on the available Romanian nouns, another just on the part of the dataset containing nouns with only 8 forms.

\section{Results}

Unlike the verbal experiments \cite{sulea19}, we first see that our language models for the Finnish as well as the German noun corpus are not able to reach performance anywhere close to the state-of-the-art \cite{yoav17}. In fact, the Finnish is only able to produce the entire list of 30 inflected forms for only 5 nouns in the test corpus and the German one for 18. We identify two reasons for this. The first one is related to the order of magnitude greater stem variance in comparison to the size of the datasets. Specifically, the Finnish noun train set contains around 6k examples, which is close to the 7k Finish verb train set, but every noun has 30 inflected forms, unlike the 6 forms for Finnish verbs. For German, the noun dataset is even smaller than the Spanish verbal dataset (2k for 8 noun forms versus 3.5k for 6 verbal forms). Clearly, the variance is too high for the language model to learn anything about the affixation patterns. Another thing it fails to learn is that it needs to generate exactly 30 forms for Finnish and 8 for German. The second reason for the failure of the model has to do with Finnish being an agglutinative language, which means that the type of stem alternations triggered during inflection (affixation) does not lead to a character cluster in the stem to be replaced with another, close in size, rather it grows in size. 

For Romanian, we report the accuracy of generating the full sequence of inflected forms for the test set containing exclusively nouns with 8 forms and the set containing all nouns by training two separate models. As we can see, indirectly giving the language model the information that all generated examples should have 8 forms by never showing it examples with less or more forms (the Nominal\_8 models) leads to just a percentage point improvement. This is in line with the German and Finnish models either predicting too many or too little forms. The deviation is however greatly improved on the Romanian dataset, which makes us conclude that it is an issue related with size of the training since the Romanian one is around 30k. We also see that increasing the max\_length from 40 to 90 does not seem to help the system significantly and that training it for longer (from 14 to 28 epochs) leads the model to overfit to regular nouns and therefore generate poorer quality inflection tables (generation accuracy for the 8 form model drops 2\% points from 14 to 28 epochs).

We also include the results from previous work for reference, although these models are either fully supervised or semi-supervised and use quite a lot of extensive pre-training. In contrast, our models use no supervision such as tags related to the context in which each generated form needs to appear (person, number).

\begin{table}[!h]
\centering
\caption{Results for full inflection generation and previous work}
\label{tab:3}
\begin{tabular}{lllllll}
      \hline
      Model & Supervision      &   Forms & Epochs & Language & max\_len & Accuracy \%\\   
     \hline
     Verbal \cite{sulea19} & Unsupervised & 6 & 14 & Romanian & 40 & 74.68 \\
     Verbal \cite{sulea19} & Unsupervised & 6 &28 & Romanian & 40 & \textbf{84.86} \\
     Nominal\_all       & Unsupervised & varies, 8 on avg. &14 &  Romanian & 40 & 71.5\\
     Nominal\_all       & Unsupervised & varies, 8 on avg.& 14 &  Romanian & 90 & 71.8\\
     Nominal\_8         & Unsupervised & 8 & 14 &  Romanian & 40 & 72.6 \\
     Nominal\_8         & Unsupervised & 8 & 14 &  Romanian & 90 & 72.2 \\
     Nominal\_8         & Unsupervised & 8 & 28 &  Romanian & 90 & 70.6 \\

     Verbal \cite{zhou_vae}    & Semi-Supervised & - & - & Romanian & - & 78.6\\
     Verbal \cite{sulea19}     & Unsupervised & 6 & 14 & Finnish & 40 & 95.50 \\
     Verbal \cite{yoav17}      & Seq2Seq & 6 & - & Finnish & - & 98.07 \\
     Verbal \cite{sulea19}     & Unsupervised & 6 & 14 & Spanish & 40 & 92.00\\
     Verbal \cite{yoav17}      & Seq2Seq & 6 & - & Spanish & - & 99.81 \\

      \hline
      
      \hline
\end{tabular}
\end{table}

In another set of experiments, we pre-trained textgenrnn on the Romanian EuroParl corpus for 14 epochs and then fine-tuned the model for another 14 epochs using our Romanian inflectional training set described above. We then proceeded to generate inflected forms by calling the \emph{generate} function with the prefix parameter set to the uninflected infinitive forms from the test set. We observed that no inflected form was generated in this setting. Instead, the model would generate chunks of text similar in style and word choice to the pre-training EuroParl corpus, suggesting that the fine-tuning corpus was too different in style and too small to impact the pre-trained model in any way. Below we reproduce one such example, where the value for the \emph{prefix} parameter was inputed as \emph{abandona} (to abandon).

\begin{center}
abandon, aderarea lucrătorilor în cadrul negocierilor comunitare care au fost adoptate în mod eficient și la noul acord care constituie o soluție pentru produsele alimentare și în același timp este deosebit de important să se prezinte o politică externă şi de aplicare a programelor de 
\end{center}

\section{Conclusions}
In this paper, we've shown that the method introduced in \cite{sulea19} of generating in a fully unsupervised manner full inflections using neural language models implemented in an off-the-shelf artificial deep recurrent neural network architecture works well in the nominal domain of morphologically rich languages. which for Romanian and Finnish is larger. in order to generate full inflections for verbs in the morphologically-rich Romanian, Spanish, and Finnish, introducing the task of \textit{unsupervised inflection generation}. We showed that even when the training dataset is small for deep learning standards and without any supervision, we can achieve accuracy close to the state of the art for Finish and Spanish and we surpass previous state-of-the-art for Romanian. 

Our generation results with the pre-trained EuroParl models also suggest that if the fine-tuning corpus is substantially different (in number of examples and structure) from the pre-training corpus (EuroParl vs.  inflection lists), the generated text will maintain the structure of the larger corpus, used in initial training. This goes against the current assumption in transfer learning which maintains that language models are unsupervised multi-task learners which can be successfully used for a new task with minimal need for fine-tuning \cite{gpt2}.

\section*{Acknowledgements}
SY would like to thank Noisebridge Hackerspace in San Francisco for use of their computing facilities.

\bibliographystyle{splncs03}
\bibliography{163}

\end{document}